# From Seed AI to Technological Singularity via Recursively Self-Improving Software

**Roman V. Yampolskiy**

Computer Engineering and Computer Science
Speed School of Engineering
University of Louisville
roman.yampolskiy@louisville.edu

**Abstract**
Software capable of improving itself has been a dream of computer scientists since the inception of the field. In this work we provide definitions for Recursively Self-Improving software, survey different types of self-improving software, review the relevant literature, analyze limits on computation restricting recursive self-improvement and introduce RSI Convergence Theory which aims to predict general behavior of RSI systems. Finally, we address security implications from self-improving intelligent software.

**1. Introduction**
Since the early days of computer science, visionaries in the field anticipated creation of a self-improving intelligent system, frequently as an easier pathway to creation of true artificial intelligence. As early as 1950 Alan Turing wrote: "Instead of trying to produce a programme to simulate the adult mind, why not rather try to produce one which simulates the child's? If this were then subjected to an appropriate course of education one would obtain the adult brain. Presumably the child-brain is something like a notebook as one buys from the stationers. Rather little mechanism, and lots of blank sheets... Our hope is that there is so little mechanism in the child-brain that something like it can be easily programmed. The amount of work in the education we can assume, as a first approximation, to be much the same as for the human child" [1].

Turing's approach to creation of artificial (super)intelligence was echoed by I.J. Good, Marvin Minsky and John von Neumann, all three of whom published on it (interestingly in the same year, 1966): Good - "Let an ultraintelligent machine be defined as a machine that can far surpass all the intellectual activities of any man however clever. Since the design of machines is one of these intellectual activities, an ultraintelligent machine could design even better machines; there would then unquestionably be an 'intelligence explosion,' and the intelligence of man would be left far behind. Thus the first ultraintelligent machine is the last invention that man need ever make" [2]. Minsky - "Once we have devised programs with a genuine capacity for self-improvement a rapid evolutionary process will begin. As the machine improves both itself and its model of itself, we shall begin to see all the phenomena associated with the terms

"consciousness," "intuition" and "intelligence" itself. It is hard to say how close we are to this threshold, but once it is crossed the world will not be the same" [3]. Von Neumann - "There is thus this completely decisive property of complexity, that there exists a critical size below which the process of synthesis is degenerative, but above which the phenomenon of synthesis, if properly arranged, can become explosive, in other words, where syntheses of automata can proceed in such a manner that each automaton will produce other automata which are more complex and of higher potentialities than itself" [4]. Similar types of arguments are still being made today by modern researchers and the area of RSI research continues to grow in popularity [5-7], though some [8] have argued that recursive self-improvement process requires hyperhuman capability to "get the ball rolling", a kind of "Catch 22" .

Intuitively most of us have some understanding of what it means for a software system to be self-improving, however we believe it is important to precisely define such notions and to systematically investigate different types of self-improving software. First we need to define the notion of improvement. We can talk about improved efficiency – solving same problems faster or with less need for computational resources (such as memory). We can also measure improvement in error rates or finding closer approximations to optimal solutions, as long as our algorithm is functionally equivalent from generation to generation. Efficiency improvements can be classified as either producing a linear improvement as between different algorithms in the same complexity class (ex. NP), or as producing a fundamental improvement as between different complexity classes (ex. P vs NP) [9]. It is also very important to remember that complexity class notation (Big-O) may hide significant constant factors which while ignorable theoretically may change relative order of efficiency in practical applications of algorithms.

This type of analysis works well for algorithms designed to accomplish a particular task, but doesn't work well for general purpose intelligent software as an improvement in one area may go together with decreased performance in another domain. This makes it hard to claim that the updated version of the software is indeed an improvement. Mainly, the major improvement we want from self-improving intelligent software is higher degree of intelligence which can be approximated via machine friendly IQ tests [10] with a significant G-factor correlation.

A particular type of self-improvement known as Recursive Self-Improvement (RSI) is fundamentally different as it requires that the system not only get better with time, but that it gets better at getting better. A truly RSI system is theorized not to be subject to diminishing returns, but would instead continue making significant improvements and such improvements would become more substantial with time. Consequently, an RSI system would be capable of open ended self-improvement. As a result, it is possible that unlike with standard self-improvement, in RSI systems from generation-to-generation most source code comprising the system will be replaced by different code. This brings up the question of what "self" refers to in this context. If it is not the source code comprising the agent then what is it? Perhaps we can redefine RSI as Recursive Source-code Improvement (RSI) to avoid dealing with this philosophical problem. Instead of trying to improve itself such a system is trying to create a different system which is better at achieving same goals as the original system. In the most general case it is trying to create an even smarter artificial intelligence.

In this paper we will attempt to define the notion of self-improvement in software, survey possible types of self-improvement, analyze behavior of self-improving software, and discuss limits to such processes.

## 2. Taxonomy of Types of Self-Improvement

Self-improving software can be classified by the degree of self-modification it entails. In general we distinguish three levels of improvement – modification, improvement (weak self-improvement) and recursive improvement (strong self-improvement).

*Self-Modification* does not produce improvement and is typically employed for code obfuscation to protect software from being reverse engineered or to disguise self-replicating computer viruses from detection software. While a number of obfuscation techniques are known to exist [11], ex. self-modifying code [12], polymorphic code, metamorphic code, diversion code [13], none of them are intended to modify the underlying algorithm. The sole purpose of such approaches is to modify how the source code looks to those trying to understand the software in questions and what it does [14].

*Self-Improvement* or Self-adaptation [15] is a desirable property of many types of software products [16] and typically allows for some optimization or customization of the product to the environment and users it is deployed with. Common examples of such software include evolutionary algorithms such as Genetic Algorithms [17-22] or Genetic Programming which optimize software parameters with respect to some well understood fitness function and perhaps work over some highly modular programming language to assure that all modifications result in software which can be compiled and evaluated. The system may try to optimize its components by creating internal tournaments between candidate solutions. Omohundro proposed the concept of efficiency drives in self-improving software [23]. Because of one of such drives, balance drive, self-improving systems will tend to balance the allocation of resources between their different subsystems. If the system is not balanced overall performance of the system could be increased by shifting resources from subsystems with small marginal improvement to those with larger marginal increase [23]. While performance of the software as a result of such optimization may be improved the overall algorithm is unlikely to be modified to a fundamentally more capable one.

Additionally, the law of diminishing returns quickly sets in and after an initial significant improvement phase, characterized by discovery of "low-hanging fruit", future improvements are likely to be less frequent and less significant, producing a Bell curve of valuable changes. Metareasoning, metalearning, learning to learn, and lifelong learning are terms which are often used in the machine learning literature to indicate self-modifying learning algorithms or the process of selecting an algorithm which will perform best in a particular problem domain [24]. Yudkowsky calls such process *non-recursive optimization* – a situation in which one component of the system does the optimization and another component is getting optimized [25].

In the field of complex dynamic systems, aka chaos theory, positive feedback systems are well known to always end up in what is known as an *attractor*- a region within system's state space that the system can't escape from [26]. A good example of such attractor convergence is the process of Metacompilation or Supercompilation [27] in which a program designed to take source code written by a human programmer and to optimize it for speed is applied to its own source code. It will likely produce a more efficient compiler on the first application perhaps by 20%, on the second application by 3%, and after a few more recursive iterations converge to a fixed point of zero improvement [26].

*Recursive Self-Improvement* is the only type of improvement which has potential to completely replace the original algorithm with a completely different approach and more importantly to do so multiple times. At each stage newly created software should be better at optimizing future version of the software compared to the original algorithm. As of the time of this writing it is a purely theoretical concept with no working RSI software known to exist. However, as many have predicted that such software might become a reality in the 21$^{st}$ century it is important to provide some analysis of properties such software would exhibit.

Self-modifying and self-improving software systems are already well understood and are quite common. Consequently, we will concentrate exclusively on RSI systems. In practice performance of almost any system can be trivially improved by allocation of additional computational resources such as more memory, higher sensor resolution, faster processor or greater network bandwidth for access to information. This linear scaling doesn't fit the definition of recursive-improvement as the system doesn't become better at improving itself. To fit the definition the system would have to engineer a faster type of memory not just purchase more memory units of the type it already has access to. In general hardware improvements are likely to speed up the system, while software improvements (novel algorithms) are necessary for achievement of meta-improvements.

It is believed that AI systems will have a number of advantages over human programmers making it possible for them to succeed where we have so far failed. Such advantages include [28]: longer work spans (no breaks, sleep, vocation, etc.), omniscience (expert level knowledge in all fields of science, absorbed knowledge of all published works), superior computational resources (brain vs processor, human memory vs RAM), communication speed (neurons vs wires), increased serial depth (ability to perform sequential operations in access of about a 100 human brain can manage), duplicability (intelligent software can be instantaneously copied), editability (source code unlike DNA can be quickly modified), goal coordination (AI copies can work towards a common goal without much overhead), improved rationality (AIs are likely to be free from human cognitive biases) [29], new sensory modalities (native sensory hardware for source code), blending over of deliberative and automatic processes (management of computational resources over multiple tasks), introspective perception and manipulation (ability to analyze low level hardware, ex. individual neurons), addition of hardware (ability to add new memory, sensors, etc.), advanced communication (ability to share underlying cognitive representations for memories and skills) [30].

Chalmers [31] uses logic and mathematical induction to show that if an $AI_0$ system is capable of producing only slightly more capable $AI_1$ system generalization of that process leads to superintelligent performance in $AI_n$ after n generations. He articulates, that his proof assumes that the *proportionality thesis,* which states that increases in intelligence lead to proportionate increases in the capacity to design future generations of AIs, is true.

Nivel et al. proposed formalization of RSI systems as autocatalytic sets – collections of entities comprised of elements, each of which can be created by other elements in the set making it possible for the set to self-maintain and update itself. They also list properties of a system which make it purposeful, goal-oriented and self-organizing, particularly: *reflectivity* – ability to analyze and rewrite its own structure; *autonomy* – being free from influence by system's original designers (*bounded autonomy* – is a property of a system with elements which are not subject to self-modification); *endogeny* – an autocatalytic ability [32]. Nivel and Thorisson also attempt to operationalize autonomy by the concept of *self-programming* which they insist has to be done in an experimental way instead of a theoretical way (via proofs of correctness) since it is the only tractable approach [33].

Yudkowsky writes prolifically about recursive self-improving processes and suggests that introduction of certain concepts might be beneficial to the discussion, specifically he proposes use of terms - Cascades, Cycles and Insight which he defines as: Cascades – when one development leads to another; Cycles – repeatable cascade in which one optimization leads to another which in turn benefits the original optimization; Insight – new information which greatly increases one's optimization ability [34]. Yudkowsky also suggests that the goodness and number of opportunities in the space of solutions be known as *Optimization Slope* while *optimization resources* and *optimization efficiency* refer to how much of computational resources an agent has access to and how efficiently the agent utilizes said resources. An agent engaging in an *optimization process* and able to hit non-trivial targets in large search space [35] is described as having significant optimization power [25].

RSI software could be classified based on the number of improvements it is capable of achieving. The most trivial case is the system capable of undergoing a single fundamental improvement. The hope is that truly RSI software will be capable of many such improvements, but the question remains open regarding the possibility of an infinite number of recursive-improvements. It is possible that some upper bound on improvements exists limiting any RSI software to a finite number of desirable and significant rewrites. Critics explain failure of scientists, to date, to achieve a sustained RSI process by saying that RSI researchers have fallen victims of the *bootstrap fallacy* [36].

Another axis on which RSI systems can be classified has to do with how improvements are discovered. Two fundamentally different approaches are understood to exist. The first one is a brute force based approach [37] which utilizes Levin (Universal [38]) Search [39]. The idea is to consider all possible strings of source code up to some size limit and to select the one which can be proven to provide improvements. While theoretically optimal and guaranteed to find superior solution if one exists this method is not computationally feasible in practice. Some variants of this approach to self-improvement, known as Gödel Machines [40-45], Optimal Ordered Problem Solver (OOPS) [46] and Incremental Self-Improvers [47, 48], have been thoroughly analyzed by Schmidhuber and his co-authors. Second approach assumes that the system has a certain level of scientific competence and uses it to engineer and test its own replacement. Whether a system of any capability can intentionally invent a more capable and so a more complex system remains as the fundamental open problem of RSI research.

Finally, we can consider a hybrid RSI system which includes both an artificially intelligent program and a human scientist. Mixed human-AI teams have been very successful in many domains such as chess or theorem proving. It would be surprising if having a combination of natural and artificial intelligence did not provide an advantage in designing new AI systems or enhancing biological intelligence. We are currently experiencing a limited version of this approach with human computer scientists developing progressively better versions of AI software (while utilizing continuously improving software tools), but since the scientists themselves remain unenhanced we can't really talk about self-improvement. This type of RSI can be classified as Indirect recursive improvement as opposed to Direct RSI in which the system itself is responsible for all modifications. Other types of Indirect RSI may be based on collaboration between multiple artificial systems instead of AI and human teams [49].

In addition to classification with respect to types of RSI we can also evaluate systems as to certain binary properties. For example: We may be interested only in systems which are guaranteed not to decrease in intelligence, even temporarily, during the improvement process. This may not be possible if the intelligence design landscape contains local maxima points.

Another property of any RSI system we are interested in understanding better is necessity of unchanging source code segments. In other words must an RSI system be able to modify any part of its source code or are certain portions of the system (encoded goals, verification module) must remain unchanged from generation to generation. Such portions would be akin to ultra-conserved elements or conserved sequences of DNA [50, 51] found among multiple related species. This question is particularly important for the goal preservation in self-improving intelligent software, as we want to make sure that future generations of the system are motivated to work on the same problem [31]. As AI goes through the RSI process and becomes smarter and more rational it is likely to engage in a de-biasing process removing any constraints we programmed into it [8]. Ideally we would want to be able to prove that even after recursive self-improvement our algorithm maintains the same goals as the original. Proofs of safety or correctness for the algorithm only apply to particular source code and would need to be rewritten and re-proven if the code is modified, which happens in RSI software many times. But we suspect that re-proving slightly modified code may be easier compared to having to prove safety of a completely novel piece of code.

We are also interested in understanding if RSI process can take place in an isolated (leakproofed [52]) system or if interaction with external environment, internet, people, other AI agents is necessary. Perhaps access to external information can be used to mediate speed of RSI process. This also has significant implications on safety mechanisms we can employ while experimenting with early RSI systems [53-61]. Finally, it needs to be investigated if the whole RSI process can be paused at any point and for any specific duration of time in order to limit any negative impact from potential intelligence explosion. Ideally we would like to be able to program our Seed AI to RSI until it reaches certain level of intelligence, pause and wait for further instructions.

**On the Limits of Recursively Self-Improving Artificially Intelligent Systems**
The mere possibility of recursively self-improving software remains unproven. In this section we present a number of arguments against such phenomenon.

First of all, any implemented software system relies on hardware for memory, communication and information processing needs even if we assume that it will take a non-Von Neumann (quantum) architecture to run such software. This creates strict theoretical limits to computation, which despite hardware advances predicted by Moore's law will not be overcome by any future hardware paradigm. Bremermann [62], Bekenstein [63], Lloyd [64], Anders [65], Aaronson [66], Shannon [67], Krauss [68], and many others have investigated ultimate limits to computation in terms of speed, communication and energy consumption with respect to such factors as speed of light, quantum noise, and gravitational constant. Some research has also been done on establishing ultimate limits for enhancing human brain's intelligence [69]. While their specific numerical findings are outside of the scope of this work, one thing is indisputable: there are ultimate physical limits to computation. Since more complex systems have greater number of components and require more matter, even if individual parts are designed at nanoscale, we can conclude that just like matter and energy are directly related [70] and matter and information ("it from bit") [71] so is matter and intelligence. While we are obviously far away from hitting any limits imposed by availability of matter in the universe for construction of our supercomputers it is a definite theoretical upper limit on achievable intelligence even under the multiverse hypothesis.

In addition to limitations endemic to hardware, software-related limitations may present even bigger obstacles for RSI systems. Intelligence is not measured as a standalone value but with respect to the problems it allows to solve. For many problems such as playing checkers [72] it is possible to completely solve the problem (provide an optimal solution after considering all possible options) after which no additional performance improvement would be possible [73]. Other problems are known to be unsolvable regardless of level of intelligence applied to them [74]. Assuming separation of complexity classes (such as P vs NP) holds [9], it becomes obvious that certain classes of problems will always remain only approximately solvable and any improvements in solutions will come from additional hardware resources not higher intelligence.

Wiedermann argues that cognitive systems form an infinite hierarchy and from a computational point of view human-level intelligence is upper-bounded by the $\Sigma_2$ class of the Arithmetic Hierarchy [75]. Because many real world problems are computationally infeasible for any non-trivial inputs even an AI which achieves human level performance is unlikely to progress towards higher levels of the cognitive hierarchy. So while theoretically machines with super-Turing computational power are possible, in practice they are not implementable as the non-computable information needed for their function is just that – not computable. Consequently Wiedermann states that while machines of the future will be able to solve problems, solvable by humans, much faster and more reliably they will still be limited by computational limits found in upper levels of the Arithmetic Hierarchy [75, 76].

Mahoney attempts to formalize what it means for a program to have a goal $G$ and to self-improve with respect to being able to reach said goal under constraint of time, $t$ [77]. Mahoney defines a goal as a function $G: N \rightarrow R$ mapping natural numbers $N$ to real numbers $R$. Given a universal Turing machine $L$, Mahoney defines $P(t)$ to mean the positive natural number encoded by output of the program $P$ with input t running on $L$ after $t$ time steps, or 0 if $P$ has not halted after $t$ steps. Mahoney's representation says that $P$ has goal $G$ at time $t$ if and only if there exists $t' > t$ such that $G(P(t')) > G(P(t))$ and for all $t' > t$, $G(P(t')) \geq G(P(t))$. If $P$ has a goal $G$, then $G(P(t))$ is a monotonically increasing function of $t$ with no maximum for $t > C$. $Q$ improves on $P$ with respect to goal $G$ if and only if all of the following condition are true: $P$ and $Q$ have goal $Q$. $\exists t$, $G(Q(t)) > G(P(t))$ and $\sim\exists t$, $t' > t$, $G(Q(t)) > G(P(t))$ [77]. Mahoney then defines an improving sequence with respect to $G$ as an infinite sequence of program $P_1, P_2, P_3, \ldots$ such that for $\forall i, i > 0$, $P_{i+1}$ improves $P_i$ with respect to $G$. Without the loss of generality Mahoney extends the definition to include the value *-1* to be an acceptable input, so *P(-1)* outputs appropriately encoded software. He finally defines $P_1$ as an RSI program with respect to *G iff $P_i(-1) = P_{i+1}$* for all $i > 0$ and the sequence $P_i$, $i = 1, 2, 3 \ldots$ is an improving sequence with respect to goal *G* [77]. Mahoney also analyzes complexity of RSI software and presents a proof demonstrating that the algorithmic complexity of $P_n$ (the *n*th iteration of an RSI program) is not greater than $O(\log n)$ implying a very limited amount of knowledge gain would be possible in practice despite theoretical possibility of RSI systems [77]. Yudkowsky also considers possibility of receiving only logarithmic returns on cognitive reinvestment: $log(n) + log(log(n)) + \ldots$ in each recursive cycle [25].

Other limitations may be unique to the proposed self-improvement approach. For example Levin type search through the program space will face problems related to Rice's theorem [78] which states that for any arbitrarily chosen program it is impossible to test if it has any non-trivial property such as being very intelligent. This testing is of course necessary to evaluate redesigned code. Also, universal search over the space of mind designs which will not be computationally possible due to the No Free Lunch theorems [79] as we have no information to reduce the size of the search space [80]. Other difficulties related to testing remain even if we are not taking about arbitrarily chosen programs but about those we have designed with a specific goal in mind and which consequently avoid problems with Rice's theorem. One such difficulty is determining if something is an improvement. We can call this obstacle – "multidimensionality of optimization". No change is strictly an improvement; it is always a tradeoff between gain in some areas and loss in others. For example, how do we evaluate and compare two software systems one of which is better at chess and the other at poker? Assuming the goal is increased intelligence over the distribution of all potential environments the system would have to figure out how to test intelligence at levels above its own a problem which remains unsolved. In general the science of testing for intelligence above level achievable by naturally occurring humans (IQ < 200) is in its infancy. De Garis raises a problem of evaluating quality of changes made to the top level structures responsible for determining the RSI's functioning, structures which are not judged by any higher level modules and so present a fundamental difficulty in accessing their performance [81].

Other obstacles to RSI have also been suggested in the literature. Löb's theorem states that a mathematical system can't assert its own soundness without becoming inconsistent [82], meaning a sufficiently expressive formal system can't know that everything it proves to be true is actually so [82]. Such ability is necessary to verify that modified versions of the program are still consistent with its original goal of getting smarter. Another obstacle, called *procrastination paradox* will also prevent the system from making modifications to its code since the system will find itself in a state in which a change made immediately is as desirable and likely as the same change made later [83, 84]. Since postponing making the change carries no negative implications and may actually be safe this may result in an infinite delay of actual implementation of provably desirable changes.

Similarly, Bolander raises some problems inherent in logical reasoning with self-reference, namely, self-contradictory reasoning, exemplified by the Knower Paradox of the form - "This sentence is false" [85]. Orseau and Ring introduce what they call "Simpleton Gambit" a situation in which an agent will chose to modify itself towards its own detriment if presented with a high enough reward to do so [86]. Yampolskiy reviews a number of related problems in rational self-improving optimizers, above a certain capacity, and concludes, that despite opinion of many, such machines will choose to "wirehead" [87]. Chalmers [31] suggests a number of previously unanalyzed potential obstacles on the path to RSI software with *Correlation obstacle* being one of them. He describes it as a possibility that no interesting properties we would like to amplify will correspond to ability to design better software.

Yampolskiy is also concerned with accumulation of errors in software undergoing an RSI process, which is conceptually similar to accumulation of mutations in the evolutionary process experienced by biological agents. Errors (bugs) which are not detrimental to system's performance are very hard to detect and may accumulate from generation to generation building on each other until a critical mass of such errors leads to erroneous functioning of the system, mistakes in evaluating quality of the future generations of the software or a complete breakdown [88].

The self-reference aspect in self-improvement system itself also presents some serious challenges. It may be the case that the minimum complexity necessary to become RSI is higher than what the system itself is able to understand. We see such situations frequently at lower levels of intelligence, for example a squirrel doesn't have mental capacity to understand how a squirrel's brain operates. Paradoxically, as the system becomes more complex it may take exponentially more intelligence to understand itself and so a system which starts capable of complete self-analysis may lose that ability as it self-improves. Informally we can call it the Munchausen obstacle, inability of a system to lift itself by its own bootstraps. An additional problem may be that the system in question is computationally irreducible [89] and so can't simulate running its own source code. An agent cannot predict what it will think without thinking it first. A system needs 100% of its memory to model itself, which leaves no memory to record the output of the simulation. Any external memory to which the system may write becomes part of the system and so also has to be modeled. Essentially the system will face an infinite regress of self-models from which it can't escape. Alternatively, if we take a physics perspective on the issue, we can see intelligence as a computational resource (along with time and space) and so producing more of it will not be possible for the same reason why we can't make a perpetual motion device as it would violate fundamental laws of nature related to preservation of energy. Similarly it has been argued that a Turing Machine cannot output a machine of greater algorithmic complexity [90].

We can even attempt to formally prove impossibility of intentional RSI process via proof by contradiction: Let's define RSI $R_1$ as a program not capable of algorithmically solving a problem of difficulty $X$, say $X_i$. If $R_1$ modifies its source code after which it is capable of solving $X_i$ it violates our original assumption that $R_1$ is not capable of solving $X_i$ since any introduced modification could be a part of the solution process, so we have a contradiction of our original assumption, and $R_1$ can't produce any modification which would allow it to solve $X_i$, which was to be shown. Informally, if an agent can produce a more intelligent agent it would already be as capable as that new agent. Even some of our intuitive assumptions about RSI are incorrect. It seems that it should be easier to solve a problem if we already have a solution to a smaller instance of such problem [91] but in a formalized world of problems belonging to the same complexity class, re-optimization problem is proven to be as difficult as optimization itself [92-95].

**Analysis**
A number of fundamental problems remain open in the area of RSI. We still don't know the minimum intelligence necessary for commencing the RSI process, but we can speculate that it would be on par with human intelligence which we associate with universal or general intelligence [96], though in principal a sub-human level system capable of self-improvement can't be excluded [31]. One may argue that even human level capability is not enough because we already have programmers (people or their intellectual equivalence formalized as functions [97] or Human Oracles [98, 99]) who have access to their own source code (DNA), but who fail to understand how DNA (nature) works to create their intelligence. This doesn't even include additional complexity in trying to improve on existing DNA code or complicating factors presented by the impact of learning environment (nurture) on development of human intelligence. Worse yet, it is not obvious how much above human ability an AI needs to be to begin overcoming the "complexity barrier" associated with self-understanding. Today's AIs can do many things people are incapable of doing, but are not yet capable of RSI behavior.

We also don't know the minimum size of program (called Seed AI [100]) necessary to get the ball rolling. Perhaps if it turns out that such "minimal genome" is very small a brute force [37] approach might succeed in discovering it. We can assume that our Seed AI is the smartest Artificial General Intelligence known to exist [101] in the world as otherwise we can simply delegate the other AI as the seed. It is also not obvious how the source code size of RSI will change as it goes through the improvement process, in other words what is the relationship between intelligence and minimum source code size necessary to support it. In order to answer such questions it may be useful to further formalize the notion of RSI perhaps by representing such software as a Turing Machine [102] with particular inputs and outputs. If that could be successfully accomplished a new area of computational complexity analysis may become possible in which we study algorithms with dynamically changing complexity (Big-O) and address questions about how many code modification are necessary to achieve certain level of performance from the algorithm.

This of course raises the question of speed of RSI process, are we expecting it to take seconds, minutes, days, weeks, years or more (hard takeoff VS soft takeoff) for the RSI system to begin hitting limits of what is possible with respect to physical limits of computation [103]? Even in suitably constructed hardware (human baby) it takes decades of data input (education) to get to human-level performance (adult). It is also not obvious if the rate of change in intelligence would be higher for a more advanced RSI, because it is more capable, or for a "newbie" RSI because it has more low hanging fruit to collect. We would have to figure out if we are looking at improvement in absolute terms or as a percentage of system's current intelligence score.

Yudkowsky attempts to analyze most promising returns on cognitive reinvestment as he considers increasing size, speed or ability of RSI systems. He also looks at different possible rates of return and arrives at three progressively steeper trajectories for RSI improvement which he terms: "fizzle", "combust" and "explode" aka "AI go FOOM" [25]. Hall [8] similarly analyzes rates of return on cognitive investment and derives a curve equivalent to double the Moore's Law rate. Hall also suggest that an AI would be better of trading money it earns performing useful work for improved hardware or software rather than attempt to directly improve itself since it would not be competitive against more powerful optimization agents such as Intel corporation.

Fascinatingly, by analyzing properties which correlate with intelligence, Chalmers [31] is able to generalize self-improvement optimization to properties other than intelligence. We can agree that RSI software as we describe it in this work is getting better at designing software not just at being generally intelligent. Similarly other properties associated with design capacity can be increased along with capacity to design software for example capacity to design systems with sense of humor and so in addition to intelligence explosion we may face an explosion of funniness.

**RSI Convergence Theorem**
A simple thought experiment regarding RSI can allow us to arrive at a fascinating hypothesis. Regardless of the specifics behind the design of the Seed AI used to start an RSI process all such systems, attempting to achieve superintelligence, will converge to the same software architecture. We will call this intuition - *RSI Convergence Theory*. There is a number of ways in which it can happen, depending on the assumptions we make, but in all cases the outcome is the same, a practically computable agent similar to AIXI (which is an incomputable but superintelligent agent [104]).

If an upper limit to intelligence exists multiple systems will eventually reach that level, probably by taking different trajectories, and in order to increase their speed will attempt to minimize the size of their source code eventually discovering smallest program with such level of ability. It may even be the case that sufficiently smart RSIs will be able to immediately deduce such architecture from basic knowledge of physics and Kolmogorov Complexity [105]. If, however, intelligence turns out to be an unbounded property RSIs may not converge. They will also not converge if many programs with maximum intellectual ability exist and all have the same Kolmogorov complexity or if they are not general intelligences and are optimized for different environments. It is also likely that in the space of minds [35] stable attractors include sub-human and super-human intelligences with precisely human level of intelligence being a rare particular [30].

If correct, predictions of RSI convergence imply creation of what Bostrom calls a Singleton [106], a single decision making agent in control of everything. Further speculation can lead us to conclude that converged RSI systems separated by space and time even at cosmological scales can engage in acausal cooperation [107, 108] since they will realize that they are the same agent with the same architecture and so are capable of running perfect simulations of each other's future behavior. Such realization may allow converged superintelligence with completely different origins to implicitly cooperate particularly on meta-tasks. One may also argue that humanity itself is on the path which converges to the same point in the space of all possible intelligences (but is undergoing a much slower RSI process). Consequently, by observing a converged RSI architecture and properties humanity can determine its ultimate destiny, its purpose in life, its Coherent Extrapolated Volition (CEV) [109].

**Conclusions**
Recursively Self-Improving software is the ultimate form of artificial life and creation of life remains one of the great unsolved mysteries in science. More precisely, the problem of creating RSI software is really the challenge of creating a program capable of writing other programs [110], and so is an AI-Complete problem as has been demonstrated by Yampolskiy [98, 99]. AI-complete problems are by definition most difficult problems faced by AI researchers and it is likely that RSI source code will be so complex that it would be difficult or impossible to fully analyze [49]. Also, the problem is likely to be NP-Complete as even simple metareasoning and metalearning [111] problems have been shown by Conitzer and Sandholm to belong to that class. In particular they proved that allocation of deliberation time across anytime algorithms running on different problem instances is NP-Complete and a complimentary problem of dynamically allocating information gathering resources by an agent across multiple actions is NP-Hard, even if evaluating each particular action is computationally simple. Finally, they showed that the problem of deliberately choosing a limited number of deliberation or information gathering actions to disambiguate the state of the world is PSPACE Hard in general [112].

Intelligence is a computational resource and as with other physical resources (mass, speed) its behavior is probably not going to be just a typical linear extrapolation of what we are used to, if observed at high extremes (IQ > 200+). It may also be subject to fundamental limits such as the speed limit on travel of light or fundamental limits we do not yet understand or know about (unknown unknowns). In this work we reviewed a number of computational upper limits to which any successful RSI system will asymptotically strive to grow, we can note that despite existence of such upper bounds we are currently probably very far from reaching them and so still have plenty of room for improvement at the top. Consequently, any RSI achieving such significant level of enhancement, despite not creating an infinite process, will still seem like it is producing superintelligence with respect to our current state [113].

The debate regarding possibility of RSI will continue. Some will argue that while it is possible to increase processor speed, amount of available memory or sensor resolution the fundamental ability to solve problems can't be intentionally and continuously improved by the system itself. Additionally, critics may suggest that intelligence is upper bounded and only differs by speed and available info to process [114]. In fact they can point out to such maximum intelligence, be it a theoretical one, known as AIXI, an agent which given infinite computational resources will make purely rational decisions in any situation.

A resource-dependent system undergoing RSI intelligence explosion can expand and harvest matter, at the speed of light, from its origin converting the universe around it into a computronium sphere [114]. It is also very likely to try and condense all the matter it obtains into a super-dense unit of constant volume (reminiscent of the original physical singularity point which produced the Big Bang, see Omega Point [115]) to reduce internal computational costs which grow with the overall size of the system and at cosmic scales are very significant even at the speed of light. A side effect of this process would be emergence of an event horizon impenetrable to scientific theories about the future states of the underlying RSI system. In some limited way we already see this condensation process in attempts of computer chip manufacturers to pack more and more transistors into exponentially more powerful chips of same or smaller size. And so, from the Big Bang explosion of the original cosmological Singularity to the Technological Singularity in which intelligence explodes and attempts to amass all the matter in the universe back into a point of infinite density (Big Crunch) which in turn causes the next (perhaps well controlled) Big Bang, the history of the universe continues and relies on intelligence as its driver and shaper (similar ideas are becoming popular in cosmology [116-118]).

Others will say that since intelligence is the ability to find patterns in data, intelligence has no upper bounds as the number of variables comprising a pattern can always be greater and so present a more complex problem against which intelligence can be measured. It is easy to see that even if in our daily life the problems we encounter do have some maximum difficulty it is certainly not the case with theoretical examples we can derive from pure mathematics. It seems likely that the debate will not be settled until a fundamental unsurmountable obstacle to RSI process is found or a proof by existence is demonstrated. Of course the question of permitting machines to undergo RSI transformation, if it is possible, is a separate and equally challenging problem.